\documentclass{article} 
\usepackage{arxiv,times}
\usepackage{cite}
\def\pdfforpub{} 

\usepackage[utf8]{inputenc}
\usepackage{lmodern}		
\usepackage{cmap}			

\usepackage{lipsum}
\usepackage{calc}
\usepackage{xifthen}
\usepackage[]{algorithm2e}

\usepackage{hyperref}

\usepackage{amssymb}
\usepackage{amsmath}
\usepackage{multirow}
\usepackage{todonotes}
\usepackage{graphics,graphicx}
\usepackage{xspace}
\usepackage{dsfont}
\usepackage{bm}
\usepackage{mathtools}
\usepackage{hyperref}
\usepackage{verbatim}

\usepackage{color, soul}
\usepackage[utf8]{inputenc} 
\usepackage[T1]{fontenc}    
\usepackage{booktabs}       
\usepackage{amsfonts}       
\usepackage{nicefrac}       
\usepackage{microtype}      
\usepackage{doi}
\usepackage{subfig}
\usepackage{clipboard}
\usepackage{amsthm}
\usepackage{graphicx,wrapfig,lipsum}
\usepackage{verbatim}
\usepackage{xcolor}
\usepackage{mathtools}
\usepackage{tabularx}
\usepackage{siunitx}
\usepackage{float}
\newcommand{\figref}[1]{Fig.\ref{#1}}

\newcommand{\eqreff}[1]{Eq.\eqref{#1}}

\newcommand{\ui}{\mathrm i}

\hypersetup{
    colorlinks,%
    citecolor=blue,%
    filecolor=blue,%
    linkcolor=blue,%
    urlcolor=blue
}
\usepackage{amsthm}
\usepackage{soul}

\ifdefined\pdfforpub
\else
\begin{abstract}

    \end{abstract}
\fi

\author{Bruno Messias F.~Resende
\thanks{ Corresponding author \texttt{devmessias@gmail.com} }
\\
	 São Carlos Institute of Physics
	 \\
	University of São Paulo, \\
	 São Carlos, SP, Brazil \\
	\\
	\And
	Eric K.~Tokuda \\
	 São Carlos Institute of Physics
	 \\
	University of São Paulo, \\
	 São Carlos, SP, Brazil \\
\And
	Luciano da Fontoura Costa\\
	 São Carlos Institute of Physics
	 \\
	University of São Paulo, \\
	 São Carlos, SP, Brazil
}

\begin{document}

\ifdefined\pdfforpub
\title{Unraveling the graph structure of tabular data through Bayesian and spectral analysis}
\maketitle

\begin{abstract}
  In the big-data age, tabular data are being generated and analyzed everywhere. As a consequence, 
  finding and understanding the relationships between the features in these data are of great relevance. Here, to encompass these relationships, we propose a graph-based method that allows individual, group and multi-scale analyses. The method starts by mapping the  tabular data into a weighted directed graph using the Shapley additive explanations technique. With this graph of relationships,  we show that the inference of the hierarchical modular structure obtained by the Nested Stochastic Block Model (nSBM) as well as the study of the spectral space of the magnetic Laplacian can help us identify the classes of features and unravel non-trivial relationships.
As a case study, we analyzed a socioeconomic survey conducted with students in Brazil: the PeNSE survey.  The spectral embedding
 of the columns suggested that questions related to physical activities form a separate group. The application of the nSBM approach not only corroborated with that but
allowed complementary findings about the modular structure: some groups of questions showed a high adherence with the divisions qualitatively defined by the designers of the survey. As opposed to the structure obtained by the spectrum, questions from the class \textit{Safety} were partly grouped by our method in the class \textit{Drugs}. Surprisingly, by inspecting these questions, we observed that they were related to both these topics, suggesting an alternative interpretation of these questions.
These results show how our method can provide guidance for tabular data analysis as well as the design of future surveys.
\end{abstract}
\else
\fi

\section{Introduction}

At the end of the nineteenth century, Edmund Landau addressed the problem of 
how to distribute a money prize among a group of chess players using a 
table of matches\cite{landauZurRelativenWertbemessung1895}. In a table of matches, each index is associated with a player and the value of
each cell represents the result of a match between two players.
Landau proposed a method that outperformed the best approaches at the time to perform prize distributions\cite{landauZurRelativenWertbemessung1895, landauUberPreisverteilungBei1915}. Landau's method is commonly known as eigenvector centrality today and it got several developments and a myriad of applications, from fraud detection to recommendation systems group\cite{vignaSpectralRanking2019}. 

Landau's method scope and limitations would become clear with the emergence of new challenges related to the analysis of tabular data from a variety of domains\cite{ghoshComprehensiveReviewTools2018, adamsSiriusMutualInformation2021}. For instance, rather than finding the best player, one may want to identify the most important socioeconomic factors to explain the variation inside of a group of people. Rather than finding a probable coalition of players or a group of chess players with similar characteristics and performance, we may want to find a group of health factors that have a similar impact on a group of diseases\cite{levy2010consumo}.

To solve these new challenges, recent works\cite{correlatioNet2008, geneNetwork2019,ambriolaokuPotentialConfoundersAnalysis2019, adamsSiriusMutualInformation2021} have been addressing the problem of exploratory analysis of tabular data by mapping it to a graph where the columns are mapped to the vertices and edges quantify the relationships between these columns. In\cite{correlatioNet2008, geneNetwork2019, adamsSiriusMutualInformation2021}, the relationships are  modeled by non-directed edges, with edge weights determined based on the mutual information or correlation values. Also, because of of this construction, the resulting graph often has multiple disconnected components. These characteristics may make up an issue if the relationships between \textit{every} pair of vertices are of importance.
In another work\cite{ambriolaokuPotentialConfoundersAnalysis2019}, the relationships are modeled by a complete directed graph with weight values expressing the global feature importance  known as \emph{gain} \cite{ganho1987, gradBoostTutorial2013}. It has been recently shown\cite{shapConsistency2019} that this measurement can lead to inconsistent results and, in addition, it also does not allow the derivation of a local explanation, in the sense of constructing a graph from a single observation (row) or a sample.

Considering these challenges in the analysis of tabular data, we are interested in answering the following questions:  \emph{1)How can we visualize the relationships between every pair of columns?};  \emph{2)How can we identify similar columns?}; 3) \emph{How can we perform the previous analyses using only a subset of columns?}.
Motivated by these three questions, we propose a framework composed of the following stages: (a) graph representation, (b) individual analysis, (c) group analysis and (d) multilevel analysis (refer to Fig.~\ref{fig:Diagram} for a an explanative diagram).

The method starts by representing the data as a graph. A microscale analysis is evaluated by considering the notions of \emph{relevance} and \emph{similarity} between columns.  A macro-level analysis must be carried out to investigate the existence of\emph{groups} of similar columns. Guided by the analyses results, one can refine these procedures (stages (a), (b) and (c)) considering only a subset of the columns. The effective implementation of the proposed approach requires choices of the algorithms for each stage. In this work, we not just propose this general framework, but we also propose principled choices suitable for the questions we aim to address.

In the graph representation stage, the edge weights was obtained through the  SHapley Additive exPlanations (SHAP)\cite{lundbergUnifiedApproachInterpreting2017} technique.  This technique was chosen for its good properties when compared with other tools to compute the importance of a feature in machine learning task. In addition, it can supply a local interpretation for each object in the data.
 The obtained complete graph, here called \emph{interpretability graph}, is initially sparsified, with the goal of removing the weak relationships between features, extending graph analysis methods that can be applied. In particular, we use the disparity filter\cite{serranoExtractingMultiscaleBackbone2009}, which is an edge filtering method that has a good performance in preserving the backbone structure of the graph.

At a group level, we propose using two complementary approaches: spectral analysis and community detection. As well known in undirected graphs, the combinatorial Laplacian have interesting spectral properties related to community structures. Those properties allow the use of this operator in graph embedding methods\cite{spectraldatascience,vignaSpectralRanking2019}. 
These methods rely on two properties of  
the combinatorial Laplacian operator:
the existence of an orthogonal basis  and the 
fact that the eigenvalues associated with them resides in the real line.
Therefore, with our interpretability graph, which is a digraph, those two properties
are not satisfied\cite{Li_Yuan_Wu_Lu_2018}. To overpass that, instead of analyzing  the combinatorial Laplacian we study the spectral information of the magnetic Laplacian operator. The theory and the applications of this operator has been in focus in the literature recently\cite{f.deresendeCharacterizationComparisonLarge2020, fanuelDeformedLaplaciansSpectral2019, magnet2021} and one of the reasons for that is because it is a Hermitian operator even for directed graphs. Here we have used the eigenfunctions associated with the magnetic Laplacian operator to map the features of a tabular data into a toroidal space aiming at exploring the data in a more detailed manner.  
Meanwhile, the analysis of the spectral space can give us a notion about how the features are connected to the interpretability graph. To explore the community structures, we must use a method specifically built for that. Here, we translate the problem of how to divide a group of features into classes into finding the communities that the vertex related to those features belongs. To do that, we have  applied the \textbf{n}ested-\textbf{S}ochastic \textbf{B}lock \textbf{M}odel (nSBM)\cite{peixotoHierarchicalBlockStructures2014, peixotoNonparametricBayesianInference2017, nsbm2021} to infer the hierarchical community structure in our graph. The nSBM revealed hierarchical relationships between the features, enabling us to explore and unravel categories that have similar or dissimilar behaviors.  Further, we analyze the correspondence between these results and those derived from spectral information associated with the graph.

The spectral and nSBM approach gives to us a way to identify groups of features and how strong they are tied. However, it does not provide subsides to understand the features at the individual level. For example, how much two features are like to each other.To compare the similarity between any two features, our approach is based on the evaluation of the cosine similarity through  the node2vec algorithm applied to the interpretability graph. These individual level analyses can identify data leakage issues and features with a similar behavior as well as allowing other data exploration approaches to the tabular data.

To refine the analyses described above, we use the spectral information of the magnetic Laplacian to split the vertices on the interpretability graph in two components. And with each component, we can evaluate again the columns using the same previous analyses.

As an application example, we employed our method to the Brazilian National Survey of Scholar’s Health from IBGE (PeNSE)~\cite{oliveira2017characteristics}.  This periodic survey has been extensively studied across the years  to understand the socioeconomic and health profiles of the students in Brazil, such as regarding illicit and licit drugs consumption\cite{pense2014drugs, pense2014drugs2}, health issues\cite{pense2014asthma}  and sedentary behavior\cite{pense2020sedentary}.
Initially, we mapped the survey to a weighted directed graph, which was subsequently filtered. Thus, using the generated graph structure, we created meaningful visualizations of the relationships between the columns through the use of force-directed algorithms and the deformed magnetic Laplacian. Specifically, our visualization unravels a division of the survey, in particular of the survey items originally classified as ``Physical activies''.  The application of the nSBM to the interpretability graph allowed different insights. For instance, we discovered that some questions such as ``Driving behavior'' were originally aggregated to the class ``Safety'' in the design of the survey, but our method suggested that they may present stronger relationships with questions related to the use of drugs. Also, the hierarchical nature of the clustering allowed inferring that questions from ``Physical activities'' could be further subdivided into more classes in the survey.  

The reported results may inspire future works aiming at exploring the effect of interdependence or confounding features in tabular data, and also provide subsidies to improve the design of surveys.

\section{Methods}

\begin{figure}[!htb]
	\centering
	\includegraphics[width=\columnwidth]{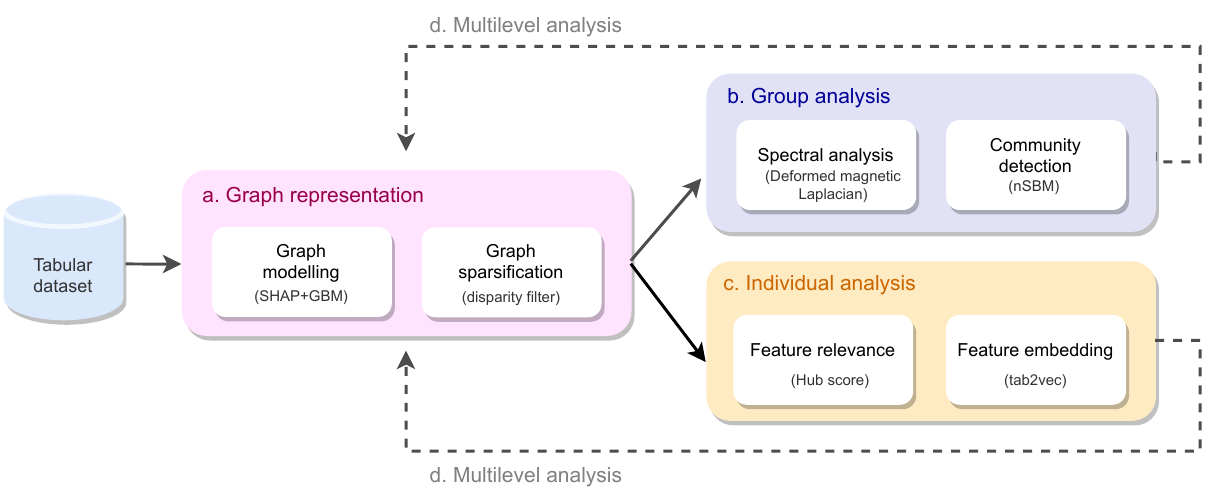}
	\caption{Flow diagram of the proposed approach. The tabular dataset is initially mapped to a weighted directed graph. The graph is reduced using a graph sparsification approach and used to perform (b) a group analysis through the use of the spectral analysis and community detection. It also allows performing (c) individual (centality measurements) and pairwise analyses (feature embeddings). These procedures can be refined (d) considering only a sample of the columns. Between parentheses, the algorithms used at each step of the evaluated framework.}
	\label{fig:Diagram}
\end{figure}

In~\figref{fig:Diagram}  we describe the method proposed in this work. We divide the method into four stages: graph representation, group, individual, and multilevel column analyses. The method starts by extracting a weighted directed graph from the tabular data, with vertices representing the columns and edges representing their relationships, weighted by the SHAP values. After that, to unravel the sparse graph representation of the tabular data, we choose to apply an edge filtering method to remove the weakest edges which finishes our pre-processing phase and allow us to employ techniques to understanding the columns and rows of the tabular data. To accomplish this understanding task, we focused on our method in the spectral information provided by the deformed magnetic Laplacian operator and the hierarchical modular structure got from the nSBM. Meanwhile, the nSBM give to us a way to group the columns into classes. The spectral information gave us a way to refine ours results based on a subset of the graph and this refinement can be repeated up to a desired granularity. Besides that those two group analyses, we also employ techniques that can give us insights about the relevance of a column (using centrality measures) or how to associate a column with a vector in a space (tab2vec). In the following, we discuss in details each step of this framework.

\subsection{Graph modelling}
A weighted directed 
graph  is a tuple $(V, E, w)$ composed by a set of vertices, $V$, a set of ordered tuples, $E$, and a weight function $w: E\mapsto \mathbb R^+$.  Each feature of the dataset associates to at least one vertex of the graph.   The directed weighted edges  
represent the relationships between two columns. 
Let $C$ be the set of columns of the tabular data.
A column $c\in C$ is randomly chosen and mapped to a set of vertices $V_c \subset V$ .
We use the remaining columns as features to train 
a gradient boosting machine (GBM) to predict the column $c$. Let $\bar V_c$ be the features columns of c.  After training, for each $v_c \in \bar V_c$, we understand the weights for each edge  ($v, v_c$) as corresponding to the
contribution of the vertex $v$ to the task of predicting the vertices related to $c$. We
 repeat this procedure for each vertex in $V$ and obtain a complete weighted directed graph.
 
A subject of particular importance concerns the \emph{contribution of $v$ to
predict $c$}. First, we want to map the tabular dataset to a graph. We require that the in-degree of vertex $v_c$ quantifies the accuracy
of the trained GBM, that is $k_{in}(v_c) = \sum_{u \in \bar V_c} w(u, v_c) = Acc(v_c)$. For instance, 
if a column has no relevant relationship with the remaining columns or cannot be explained by them, the in-degree is low, which reduces the contribution of the vertex to the overall structure of 
the graph.

This accuracy is used to calculate the weights of the edges. Let  $\epsilon(u\rightarrow v_c) \in \mathbb R^+$ be a function that quantifies the contribution of a column associated with $u$
to the task of predicting the values of column $p$ using the GBM. Here, we choose the weight function of an 
edge $(u, v)$ as:

\begin{eqnarray}
w(u, v) = Acc(v)
	\frac{
		\epsilon(u\rightarrow v)
	}{
		\sum\limits_{z\in V}\epsilon(z\rightarrow v)
	}
\label{eqVar2graph}.
\end{eqnarray}



Next we discuss how to choose $\epsilon$.
To use \eqref{eqVar2graph} and consequently, to construct the 
interpretability graph, it is necessary to choose a way to explain the
prediction of a given variable $v$ due to the
presence of a feature $u$. There is a wide range of
methods in the literature to achieve this\cite{molnarInterpretableMachineLearning}.
In this work, we opted to use the SHapley Additive exPlanations (SHAP) 
\cite{samekLearningExplainableTrees2020,lundbergLocalExplanationsGlobal2020}. The SHAP\cite{kuhnContributionsTheoryGames1953} method was motivated in the theory of cooperation games and works by quantifying the marginal contribution of a feature to a single prediction task. 

Since the SHAP value is calculated for each  element of the dataset, we have a different graph defined
by~\eqreff{eqVar2graph} for each instance. For example, if the tabular data corresponds to a survey, the graph can be used to study the answers of each person. Although this local exploration allows associating a graph with each instance in the data, in this work we focus on a single graph to describe the entire dataset. In this case, the weight of edge $(u,v)$ is defined as the mean of the
absolute values of SHAP, that is
\begin{eqnarray}
	w(u, v) = Acc(v)
	\frac{
		\mathbb E [ |\mathrm{SHAP}_i(u\rightarrow v)|]
	}{
		\sum\limits_{z\in V}\mathbb E[|\mathrm{SHAP}_i(z\rightarrow v)|]
	}
	\label{eqShapMean}.
	\end{eqnarray}

\begin{figure}[ht]
	\centering
	\includegraphics[width=.98\columnwidth]{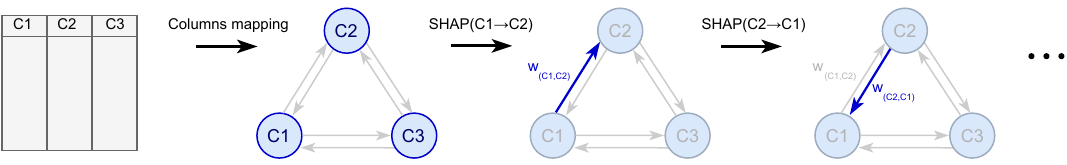}
	\caption{Construction of the interpretability graph. Initially the columns of the tabular data are mapped to a complete directed graph. Next, the weight $w_{(C1,C2)}$ of the arc $(C1,C2)$ is computed from the SHAP value of $C1$ in the prediction task of $C2$. The same procedure is repeated to assign the weight of every arc in the graph.}
	\label{fig:mapping}
\end{figure}


\subsection{Graph filtering}

The obtained interpretability graph is, by construction, complete. As a result, 
the posterior processing  may be difficult or even unfeasible. 
One of the reasons is the high computational cost associated to the processing of the entire graph. Another reason relates to the excess of information, which may end up blurring the objects of interest\cite{cosciaAtlasAspiringNetwork2021}.

A simple approach to reduce the number of edges and to enhance the interpretability of the graph visualization techniques consists in the application of
a naive threshold to the edge weights so as to keep just the strongest
connections. However, it is hard to choose and justify the value used 
for the  threshold parameter\cite{cosciaAtlasAspiringNetwork2021}. In addition,
this method can create many disconnected components. 



In the last decade, a large number of graph filtering methods (a.k.a. graph sparsification) has been developed in order to mitigate the issues present in the naive threshold-based edge filtering approach 
 \cite{serranoExtractingMultiscaleBackbone2009,marcaccioliPolyaUrnApproach2019,batsonSpectralSparsificationGraphs2013}. In this work, we adopted  
the disparity filter criterion developed by\cite{serranoExtractingMultiscaleBackbone2009}  to filter the edges.
 
Let  $s(u)=\sum\limits_{v\in V | (u, v) \in E}  w(u, v)$ be the out-degree of a 
feature associated with the node $u$ in the interpretability graph. Defined in this way, $s(u)$ 
is related to the contribution of  feature $u$ to explain the outputs of all remaining features.	
Thus $p(u ,v)={w(u, v)}/{s(u)}$ quantifies how the explanation given by the feature $u$ in the task of predicting feature $v$ contributes to the total amount of interpretability of the feature $u$. Then, with 
$k_{out}(u)$  being the out-degree of  node $u$,  we can associate with each edge $(u, v)$ the
following quantity
\begin{eqnarray}
w_\alpha(u, v) =  1 - (k_{out}(u)-1)\int\limits_{0}^{p(u, v)} (1-x)^{k_{out}(u)-2}\mathrm d x.
\end{eqnarray}

Edges with  $w_{\alpha}$ above a given threshold $\alpha \in [0, 1]$ are filtered out. Therefore, this method allows to filter the edges and at the same time keep the graph backbone, as pointed in \cite{serranoExtractingMultiscaleBackbone2009}. 


\subsection{Spectral analysis}

In the previous sections we discussed the construction of the weighted directed graph from tabular data and how to extract insights from this data structure. Here we discuss how the spectral information of the magnetic Laplacian can be used to unravel clustering of features.

The derivation of the magnetic Laplacian formalism requires decomposing the weight function
between a symmetrical  $w_s(u, v)={\big(w(u,v)+w(v, u)\big)}{/2}$  and   an asymmetrical  
$w_a(u, v)={\big(w(u,v)-w(v, u)\big)}{/2}$ components. This 
allows the definition of a flow function in each vertex $v$ due to $u$ as $a(v, u) = 2w_a(u, v)$.
With the decomposition, each digraph results in an associated undirected version
$G_s=(V, E_s, w_s)$, which relates to the Laplacian operator, $L$, by:
\begin{align}
(L f)(u) = 
f(u)d(u)
-\sum\limits_{v\in V}w_s(u, v)f(v),
\label{eqCombL}
\end{align}
where  $d(u)=\sum\limits_{v\in V}w_s(u, v)$.

As can been seen, the combinatorial Laplacian for the undirected graph is symmetric.


The second term of the right hand side of Eq.\eqref{eqCombL} needs to be modified to deal with the directionality information of the digraph. To do so, the directionality information is treated as a phase perturbation, formally
represented by a function whose domain corresponds to the edge set of the directed graph. This function has the following form:

\begin{align}
\gamma_q(u, v) = e^{2\pi \ui q a(v, u)}
\end{align}

which inserted in the second term of right-hand side~Eq.\ref{eqCombL}  gives us the magnetic Laplacian, $\mathcal L_q$, 

\begin{align}
(\mathcal L_q f)(u) &= 
f(u)d(u)
-\sum\limits_{v\in V}w_s(u, v)\gamma_q(u, v)f(v) 
\label{eqMagL}
\end{align}
where $q\in[0, 1]$ is a parameter called \emph{charge} because of historical reasons\cite{shubinDiscreteMagneticLaplacian1994a}.

It is convenient to define a normalized version of the magnetic Laplacian, $\mathcal H_q$, as

\begin{align}
(\mathcal H_q f)(u) = 
f(u)
-\frac{
	\sum\limits_{v}w_s(u, v)\gamma_q(u, v)f(v)  
}{
	d(u)
}.
\label{eqMagNormedL}
\end{align}

Noticeably, the magnetic operator can be represented by 
Hermitian matrx which is not the case of combinatorial operator
for digraphs\cite{f.deresendeCharacterizationComparisonLarge2020}. In
addition,  the magnetic Laplacian 
is a positive semi-definite operator. The positive semi-definite and hermiticity
properties of the magnetic Laplacian allow constructions of physical analogies which 
can be used to characterize digraphs\cite{f.deresendeCharacterizationComparisonLarge2020}.
In addition, the phases of a given eigenvector of the normed 
magnetic Laplacian~\eqref{eqMagNormedL}, $\mathbf v_q^{(l)}\in \mathbb
C^{|V|}$ capture the notion of circularity in the graph.
For example, the phases of the eigenvector associated with the
lowest eigenvalue of
$\mathcal H_q$
is the approximated solution for the group
synchronization problem related
with the magnetic Laplacian\cite{fanuelMagneticEigenmapsVisualization2018}. In mathematical terms this problem  searches for a mapping $\theta: V\mapsto [0, 2\pi) $ which minimizes the following function
\begin{align}
\eta_{c}(\theta) &= 
\frac{1}{2 \mathrm{vol}(G_s)}
\sum\limits_{u, v \in V}
w_s(u, v)
\left|
e^{\ui \theta(u)}-\gamma_q(u, v)e^{\ui \theta(v)}
\right|^2
\label{eqCircFrustation}
\end{align}
where $\mathrm{vol}(G_s) = \sum\limits_{u\in V} d(u)$.

The phases of the second eigenfunction
of~\eqref{eqMagNormedL} also have a remarkable
property in the  sense that this phase can approximately solve
a graph-cut problem\cite{imageSegSpectra, fanuelMagneticEigenmapsVisualization2018}. 



\subsection{Community detection} 

In principle, a class of features having similar interpretation behavior should belong to the same  community  in the proposed interpretability graph. Therefore, to understand the relationships between the features, it is first necessary to define first how these communities can be identified. One possibility to do that is to use a modularity optimization method\cite{Newman_Girvan_2004}. Unfortunately, this method has some  drawbacks. For example, 
it can find communities even in a random graph
\cite{guimera2004}. Thus, this can gives to us a meaningless division between the feature of a tabular data. Fortunately, the  non-parametric Bayesian method called
\textbf{n}ested
\textbf{S}tochastic
\textbf{B}lock \textbf{M}odel (nSBM)\cite{peixotoHierarchicalBlockStructures2014} mitigates that.

The nSBM method is the hierarchical formulation of the well-known 
Stochastic
Block Model (SBM)\cite{peixotoNonparametricBayesianInference2017, sbmTopicModelScience}. The major difference between SBM and nSBM is that 
the latter proceeds by agglomerating graph communities into levels, which
represent blocks modeled 
by a SBM. Using this hierarchical construction, nSBM overcomes some issues 
of its counterpart, such as the inefficiency in identifying small graph communities\cite{peixotoHierarchicalBlockStructures2014}. 

In essence, the SBM  performs a Bayesian inference on a set of
parameters of a generative 
graph model. Such parameters are the vertex partitions, that is, the sizes and the number 
of blocks, and the probability of connections inside and outside those
partitions. 
In mathematical terms, let $b$  be a set of vertex partitions and 
$\theta$ the parameters of a given generative model for a graph $G$, 
the Bayesian problem is given by:

\begin{eqnarray}
    P(b| G) =  \frac{P(G| b,\theta )P( b,\theta )}{P(G)},
\end{eqnarray}
where $P(G)$ it is the model evidence. 

The nSBM
uses the non-parametric framework proposed by Peixoto and it is able to efficiently infer 
 the block-hierarchical structures and, thus, to understand the modular organization of the graph. Consequently, using this method a user can unravel the relationships between features in the dataset.

\subsection{Feature embedding (tab2vec)}


We used the inferred community structure to group features (columns) on the tabular data. However, the inferred communities can not be used to define a notion of proximity between two features in the same community. Furthermore, this notion of proximity can be used to unravel features that are most similar in the interpretability graph. To define this notion, we choose to use the node2vec algorithm that allowed us to associate to each feature of the tabular data to a position in an Euclidean space, tab2vec.

\subsection{Multilevel analysis}
We can repeat all previous procedures for this subset of columns. The reason behind this step is analogous to the graph filtering stage: the density of the graph may obfuscate the identification of the features and the features interaction. While similar in motivation, this stage differs from graph filtering by considering a different set of columns to construct the graph, in contrast to the filtering of edges in the graph filtering step. This step also differs from considering the subgraph of the ``full'' interpretability graph induced by the selected vertices because here we are considering a different prediction task (without the removed features). 

To select this subset of vertices (columns) we used the magnetic eigenmaps of the interpretability graph to separate vertices as discussed in 
\cite{fanuelMagneticEigenmapsVisualization2018}. We can think that this is analogous with the techinique of using the eigenvectos of the combinatorial Laplacian to perform image segmentation\cite{imageSegSpectra}
\section{Case study: PeNSE}
The adolescence phase may strongly impact adulthood. For this reason, different surveys have focused on the related subjects~\cite{grunbaumYouthRiskBehavior2004,currie2008inequalities}.  The PeNSE (National Survey of Scholar's Health)~\cite{oliveira2017characteristics} is a survey organized by the Brazilian Institute of Geography and Statistics (IBGE), with collaboration of the Ministry of Health and of the Ministry of Education. Its mission is to better understand the risk factors and health profiles of the teenagers in Brazil.

The three editions of the survey (2009, 2012 and 2015)
targeted students regularly enrolled in a Brazilian school, public or private, at the 9th grade, which often corresponds to fourteen-year-old teenagers. This school age was chosen considering the international ethic guidelines of age to conduct socioeconomic questionnaires targeted at the teenagers group. Here we have explored the 2015 edition which inquired almost $130,000$ students in Brazil\footnote{The data is public  and available here \href{https://ftp.ibge.gov.br/pense/2015/microdados/PeNSE_2015_AMOSTRA1.zip}{https://ftp.ibge.gov.br/pense/2015/}}.

The survey consists in an electronic questionnaire comprising questions from diverse areas, such as the respondents' socioeconomic context: parents' level of education, profession, possession of goods; health, including sexual, oral and mental health; eating habits and the risk factors; family relationships and domestic violence; and the infrastructure provided by the school.

This dataset has already been explored by~\cite{levy2010consumo}, where the authors explored the association between key indicators to sociodemographical profiles. For example, a healthy nutrition indicator, which takes into account the frequency of meals and the consumption of other type of foods, was found to be associated with the age, gender and socioeconomic profile. The analysis was constrained to a linear analysis (linear regression) between these markers.  This dataset has also been explored in other works~\cite{maltaBullyingBrazilianSchools2010,maltaTrendRiskProtective2014b}, but focusing on specific set of features, such as related to bullying or chronic diseases.

\subsection{Force-directed layout and the effect of the disparity filter}

\begin{figure}[!htb]
	\centering
	\includegraphics[width=.98\columnwidth]{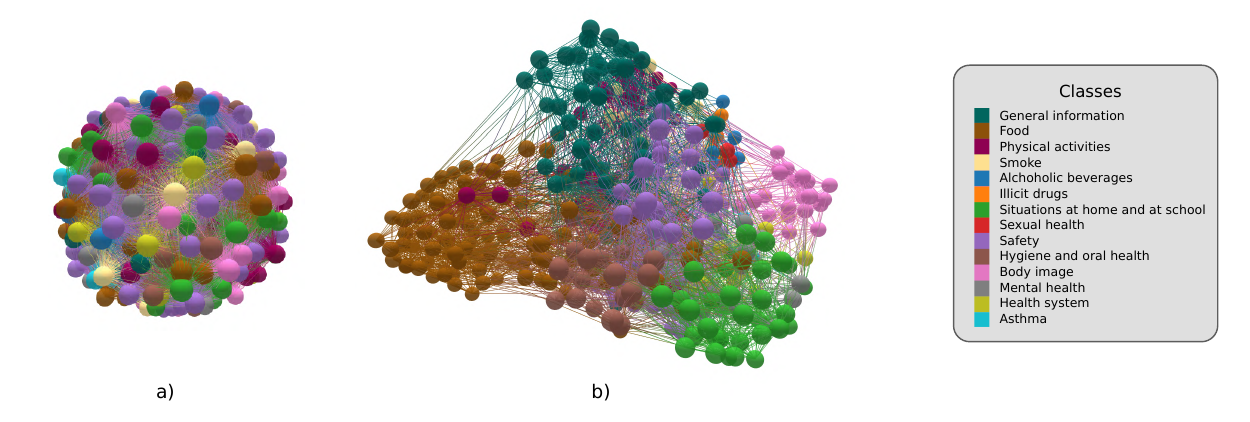}
	\caption{Interpretability graph of the PeNSE dataset. The nodes represent the features and the edges represent the relationship between pairs of features, considering our approach. In (b) the graph was initially filtered using a disparity filter, through \eqref{eqVar2graph} with parameter $0.1$. The vertices disposition is given by the force-directed algorithm. The colour of the vertices/edges corresponds to the group of each variable, provided by the dataset. 
	A good correspondence can be observed  between \emph{spatial} communities and \emph{colours} , such as the brown group at the bottom portion of the figure.}
	\label{figPENSEForceDirected}
\end{figure}

We first discuss how our method could unravel groups of questions in the PeNSE
survey. To do so, we first created the interpretability graph as previously discussed and removed less important edges using the disparity 
filter. In~\figref{figPENSEForceDirected} we show the force-directed visualization of both the complete graph (\figref{figPENSEForceDirected}(a)) as well as the sparse graph obtained after application of the edges filtering method (\figref{figPENSEForceDirected}(b)). The hairy-ball appearance of the complete graph does not allow a direct interpretation. In contrast, when the disparity filter with a $\alpha=0.1$ was applied to the complete graph, group structures starts to appear. A visual inspection shows that the questions related with physical activities seem involve two separated groups.  However, as it is well known, the force-directed embedding can be subjectively interpreted by the person who is seeing the graph. Therefore, any insight given by this method should be verified by  more formal methods. Thus, in the following we will investigate more about how this group of questions behaves in the spectral space and in the inferred modular structure.

\subsection{Spectral embedding of the PeNSE features}

\begin{figure}[ht]
	\centering
	\includegraphics[width=.5\columnwidth]{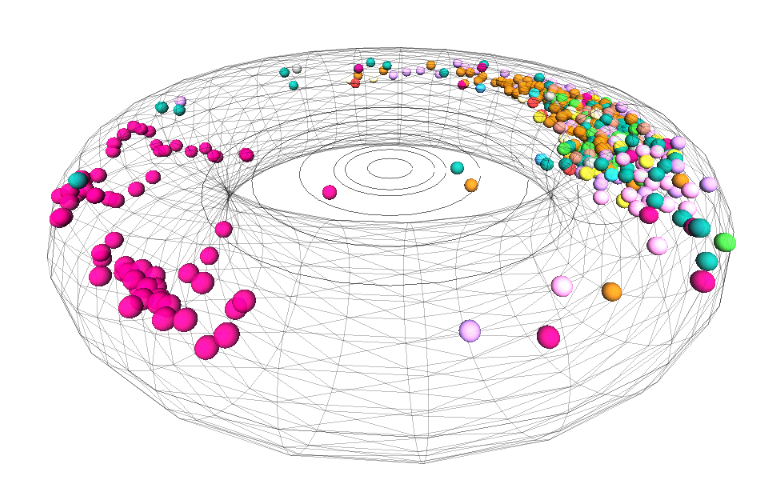}
	\caption{Spectral information embedded in a toroidal space. Features with higher hub scores are closer to the axial center. A well defined cluster of pink vertices can be observed on the left, which represent questions from class \textit{Physical Activities}. A magnetic eigenmap embedding with $q=1/10$ was utilized.}
	\label{figPENSETorus}
\end{figure}

In \figref{figPENSETorus} we present the toroidal embedding using the first two phases of
the magnetic Laplacian, with $q=1/10$ and the hub score as the radial coordinate. The highest hub score questions in the survey according are  close to the center. The embedding shows that the questions related to 
physical activities are grouped in a well separated cluster by the magnetic embedding. 
Therefore, we must expect that the questions related to physical activities
form a group more strongly related with itself. In addition, if a more detailed analysis is requested, a graph-cut approach can been done in the toroidal embedding aiming at removing most of these questions, followed by the application of our method to the reaming columns data aiming at complementing the analysis of other questions.

\subsection{Hierarchical categorization of the features}

\begin{figure}[ht]
	\centering
	\includegraphics[width=.75\columnwidth]
	{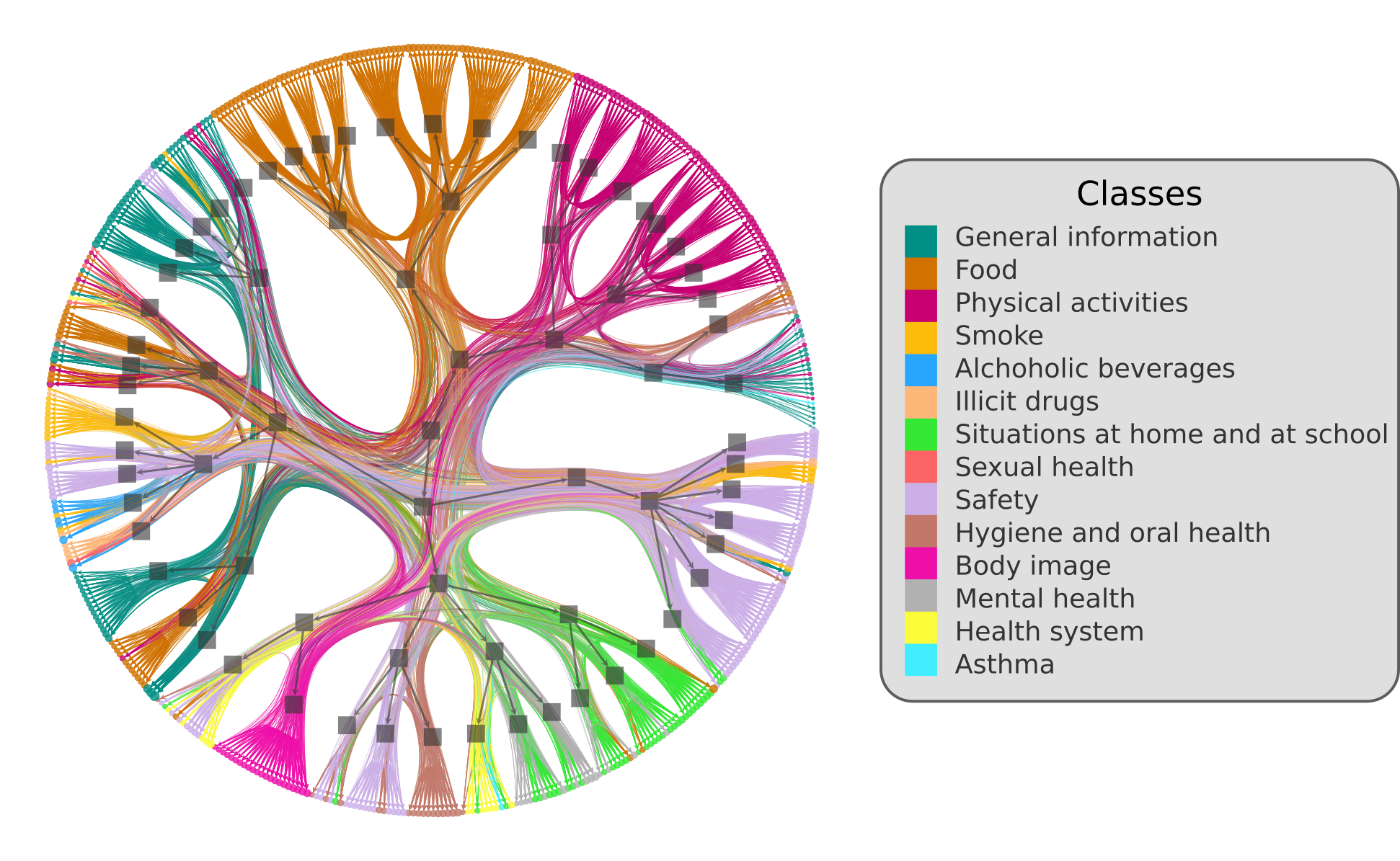}

	\caption{Circular visualization~\cite{peixotoHierarchicalBlockStructures2014} of the filtered interpretability graph with edge bundling. Vertices in the enclosing circle represent the features and the directed edges show the relationship between two features. Vertices are grouped according to the modular structure and the color of each vertex represents the class of the question in the survey. The overlaid hierarchical structure symbolizes the hierarchy of the communities.}
	\label{figPENSESBM}
\end{figure}

Community detection is generally a hard problem and this difficulty stems, in part, from the absence of a clear and common definition of what a community is~\cite{peixotoHierarchicalBlockStructures2014}. The nSBM approach attempts to mitigate this issue by proposing a statistically principled approach to identify the modular structures. In this work we used the graph-tool\footnote{https://graph-tool.skewed.de/} implementation of nSBM\cite{peixotoNonparametricBayesianInference2017, sbmNsbm2020}.  We show in~\figref{figPENSESBM} the circular visualization of the filtered interpretability graph of the features in the PeNSE survey provided by nSBM.
The directed graph with gray vertices and edges represents 
the hierarchical structure of the communities of the questions. The vertices are positioned according to the modular structure of the graph and the color of the edges and of the nodes represents the class to which each question belongs in the survey. Such classes 
were originally defined by the designers of the survey.  Thus, communities of vertices with the same color mean a correspondence between the modular structure predicted by the method and the qualitatively classification of the questions in the questionnaire.

This hierarchical circular visualization in~\figref{figPENSESBM} allows different types of analyses, but two are of particular interest regarding the analysis of the survey. The first relates to the positioning and grouping of the vertices and their correspondence with the divisions proposed in the survey. The second has to do with the connections among the areas, i.e., the existence of dominant areas to which a group of features may connect to.

\begin{figure}[ht]
	\centering
	\subfloat[]{%
		\centering
		\includegraphics[width=.25\columnwidth]
		{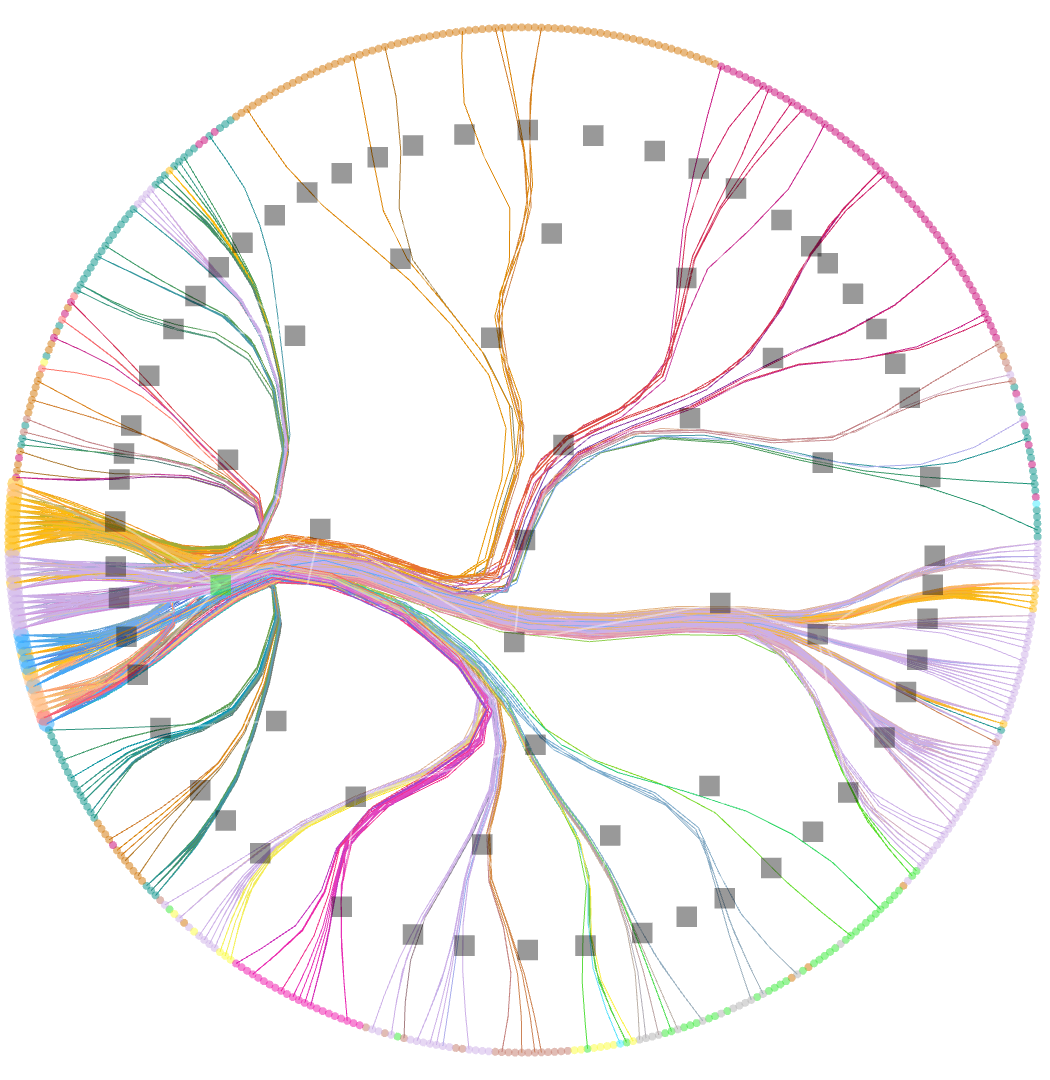}
		}
	\qquad
	\subfloat[]{
		\centering
		\includegraphics[width=.25\columnwidth]
		{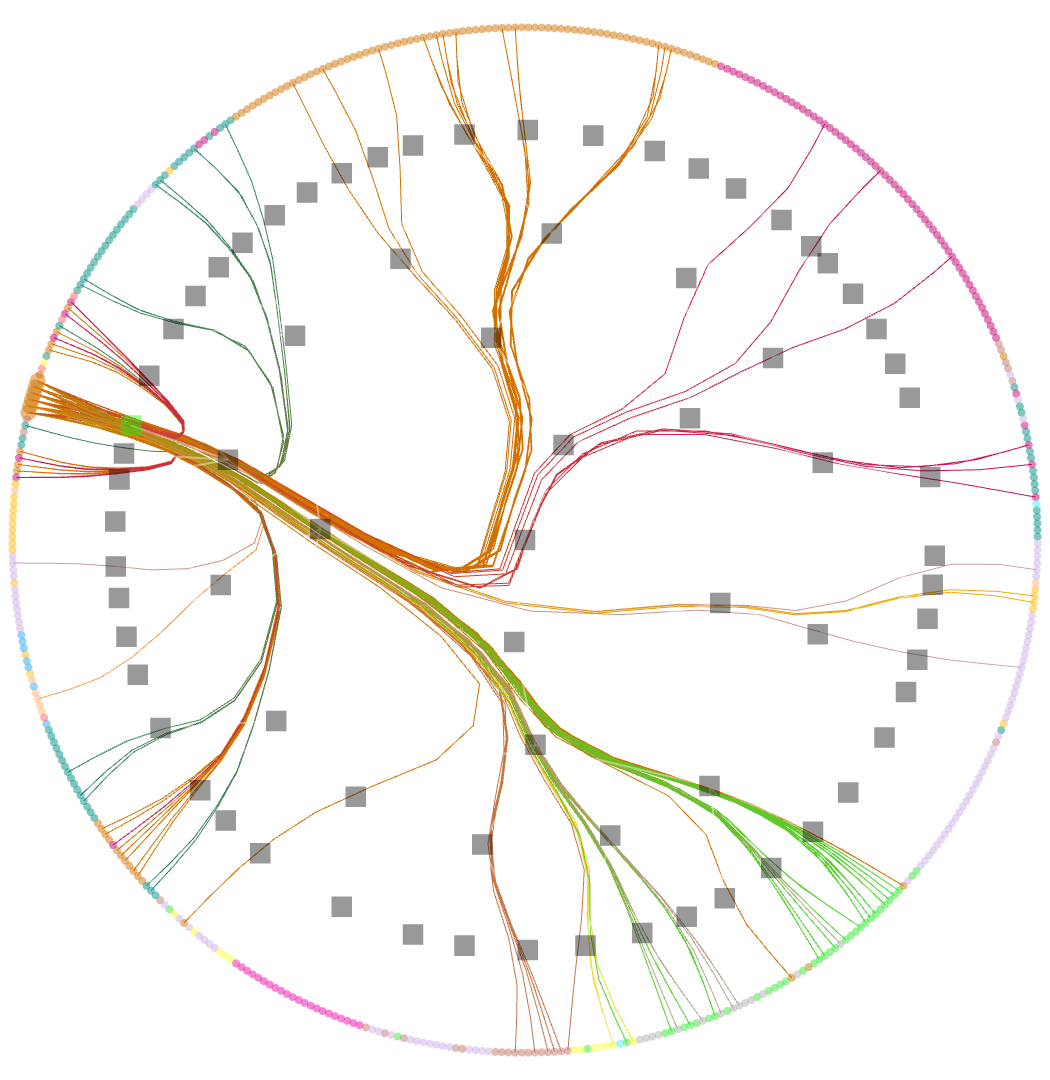}
		}
			\caption{Particular groups in the hierarchical community visualization. In (a), part of the features related to safety are located in a separate group, possibly showing a better grouping of the features. In (b), the highlighted features in orange have strong connections to green vertices, which is expected, according to the classification of the features proposed in the questionnaire.}
			\label{fighighlight}
\end{figure}

In~\figref{figPENSESBM}, one can readily see a high correspondence of the obtained grouping of the questions and the divisions of the survey for at least two classes: \textit{Food} (brown) and \textit{Body} image (magenta). Whereas the class \textit{Safety} (pink) presents a considerable agreement, part of the features were positioned by the method separately on the left region of the circle, grouped with questions related to the consumption of drugs (\figref{fighighlight}(a)). This shows that an alternative classification of the features on the left could be as pertaining to the class of \textit{Illicit drugs}. Important to emphasize that the nSBM approach is completely automatic and non-subjective, solely based on the pattern of responses in the survey.

In~\figref{fighighlight}(b), the small group in orange, on the left, is emphasized. This visualization allows us to see that this group has high connectivity to the green group, on the bottom part of the circle. The class in orange corresponds to \textit{Food} and the highlighted vertices correspond to questions related to eating with parents. The highlighted vertices in green, in turn, represent questions that deal with the relationship of the teenager and their parents. That may be understood as the strong relationship, from the point of view of the student, of a healthy relationship with the parents and sharing regular meals with them. Again, it potentially indicates another possibility of organizing these questions in the questionnaire.

\begin{figure}[ht]
	\centering
	\subfloat[]{
	\centering
	\includegraphics[width=.25\columnwidth]
	{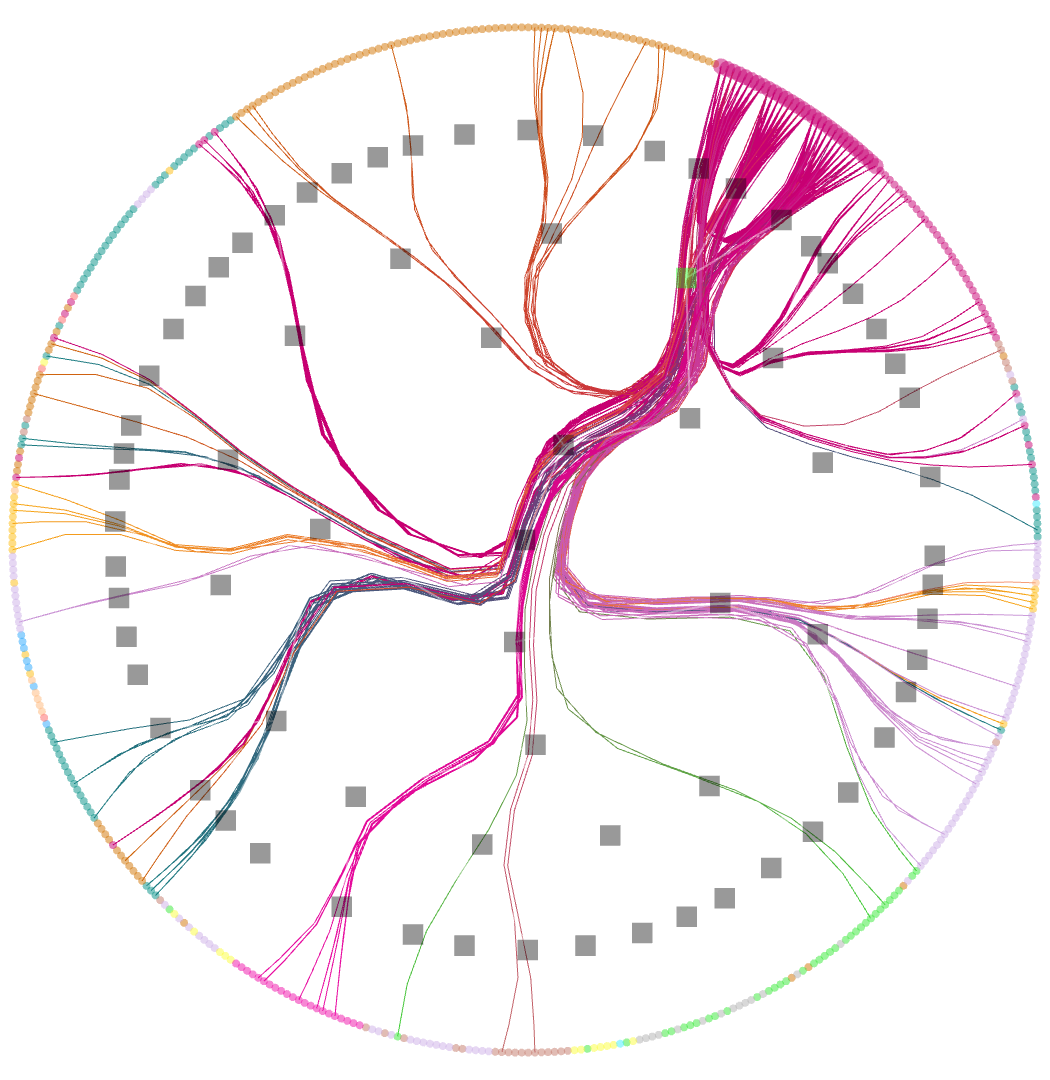}
	}
	\qquad
	\subfloat[]{%
	\centering
	\includegraphics[width=.25\columnwidth]
	{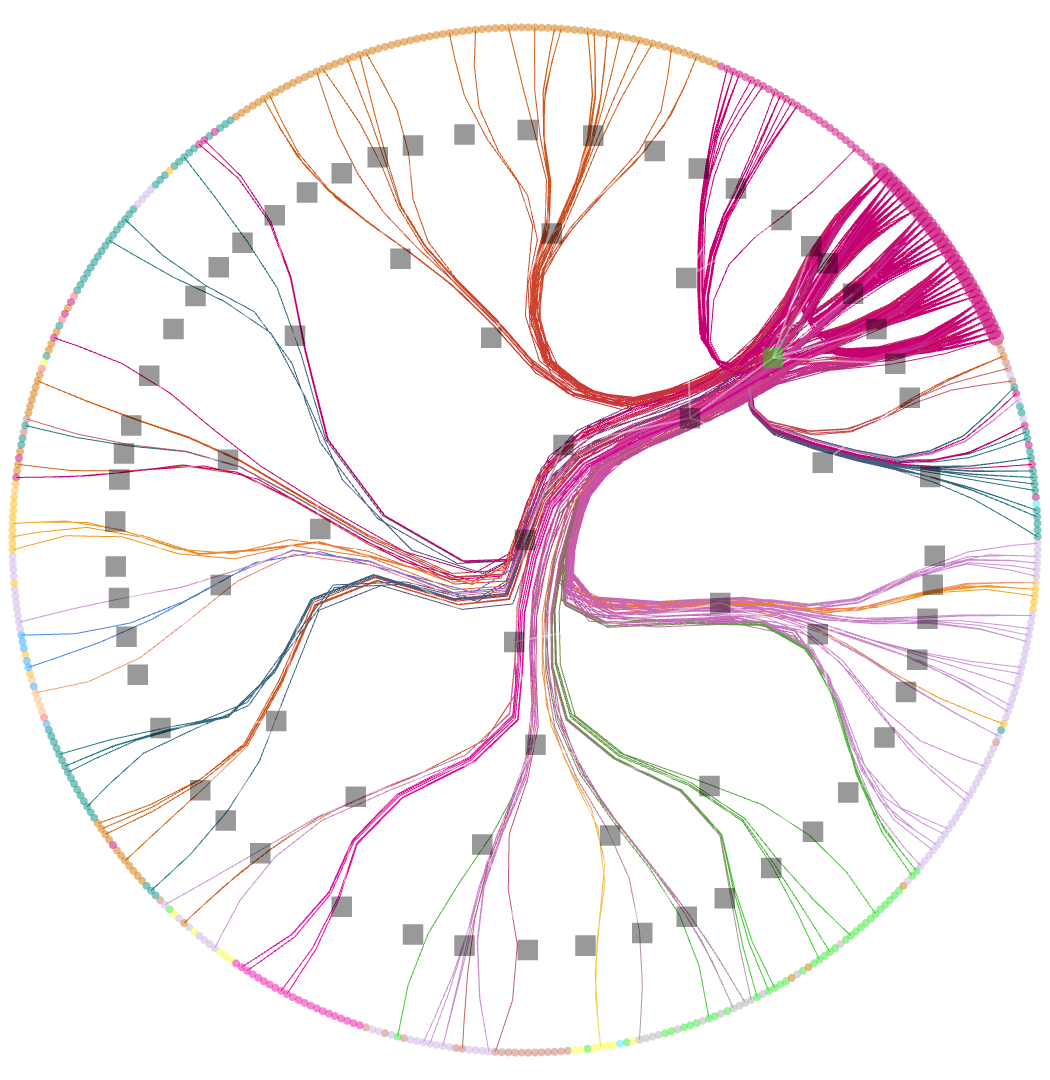}
	}
	\caption{Questions originally categorized as related to Physical Activities. While being positioned close to each other in the circle, they are divided into two groups. In (a) they are related to general sports while in (b) they are related to the activities due to their socioeconomic condition, such as commute on foot.}
			\label{figphysical}
\end{figure}

Furthermore, the hierarchical nature of 
nSBM allows a more detailed categorization of the features. 
Most of the questions related to Physical activities (in violet) are positioned in the same region of the circle, but they are grouped into two distinct subgroups (see~\figref{figphysical}). By inspecting the questions in these subgroups, we noticed that the group in~\figref{figphysical}(a) is related to entertaining activities, such as playing soccer or dancing, while the other (\figref{figphysical}(b)) relates to physical activities required by the socioeconomic condition of the respondent, such as walking or cycling from home to school (see the most relevant questions in Table~\ref{tableBicicleta}). That is related to the fact that in developing countries mobility relates to the socioeconomic level in different ways~\cite{da2008multiple}.

The proposed method groups similar questions, such as from Table~\ref{tableBicicleta}, in nearby regions in the graph. Whereas this analysis could be done manually for visualization purposes, an alternative approach is to performing it  automatic and less subjective way. For instance, the questions could be mapped into a vectorial space, \textit{tabular2vec}. In appendix~\ref{sec:tabular2vec}, we further discuss this idea and  present some results. These preliminary results seem to be consistent with our findings.

\begin{table}[ht]
	\centering
	\caption{Questions with highest hub score in the community highlighted in~\figref{figphysical}(b). It can be observed that while in fact they are related to the group ``Physical acvitivies'', they are conditioned by the socioeconomic conditions of the respondents.
 \vspace{.5em}}
	\begin{tabular}{l}
		\toprule
	``During the last 7 days, in how many days you went on foot or by bicycle to school?''\\ 
		``During the last 7 days, in how many days you came back on foot or by bicycle to school?''\\ 
		``When you go to school on foot or by bicycle, how long does it take?''\\
		``When you come back to school on foot or by bicycle, how long does it take?''\\
		\bottomrule
	\end{tabular}
	\label{tableBicicleta}
\end{table}

\subsection{Features with a similar interpretation structure as revealed by a \textit{tabular2vec} approach}
\label{sec:tabular2vec}

The hierarchical structure obtained from nSBM provides a mesoscale interpretation of the relationships between the features allowing addressing questions about how we could group different questions in a survey and how strongly those questions are tied. In addition, we could  get some clues about factors, as in the case of PeNSE survey.  However,  suppose that we are seeking for possible data-leakage issues or investigating specific factors in a tabular data.
This problem can be translated to \emph{Given a vertices, which relates to a feature, what is the set of other vertices that have the most similar structure in the interpretability graph?}  One way to answer this is to use a word embedding approach, such as node2vec\cite{node2vec2016}, applied in the context of graphs. Here, we performed a node2vec embedding using the interpretability graph. As an example, Table \ref{tableSexEducation} shows the top four questions with the highest cosine similarity to the question {``At school, have you ever received pregnancy prevention counseling?''}. Notice that the question {``At school, have you ever received advice on how to get condoms for free?''} resulted in a cosine similarity of 0.99 with the reference question, meaning that the embeddings of these two different questions are almost the same. This is reasonable considering the fact that interviews aiming counseling youngsters about pregnancy would also talk about condoms. Therefore, this suggests that we can use word embedding and the cosine similarity values in order to identify data relationship issues.

\begin{table}[ht]
	\centering
	\caption{Most similar questions according to the cosine similarity to the question 
{``At school, have you ever received pregnancy prevention counseling?''}. Questions sorted by the cosine similarity value (first column). \vspace{.5em}}
	\begin{tabular}{c|l}
		\toprule
		Cos. & Question \\\midrule
    0.99&{``At school, have you ever received advice on how to get condoms for free?''}\\
    0.98&{``At school, have you ever received advice about AIDS or other sexually transmitted diseases?''}\\
    0.87&{``Have you heard about the vaccination campaign against the HPV virus?''}\\
    0.52&{``In the last twelve months, how many times did you get involved in a fight (a physical fight)?''}\\
		\bottomrule
	\end{tabular}
	\label{tableSexEducation}
\end{table}

\subsection{Multilevel analysis}
We can do the multilevel analysis defining criteria to select a subset of the vertices sets in the interpretability graph, which represents the set of columns in the tabular data. Here, we proposed and use the magnetic eigenmaps as a tool to perform this column segmentation, similar to image segmentation in computer vision using the combinatorial Laplacian eigenvectors. We present the impact of this segmentation on embedding space got using the t-sne and UMAP technique\cite{tsneART, umapPaper}.

\begin{figure}[ht]
	\centering
	\includegraphics[width=.75\columnwidth]
	{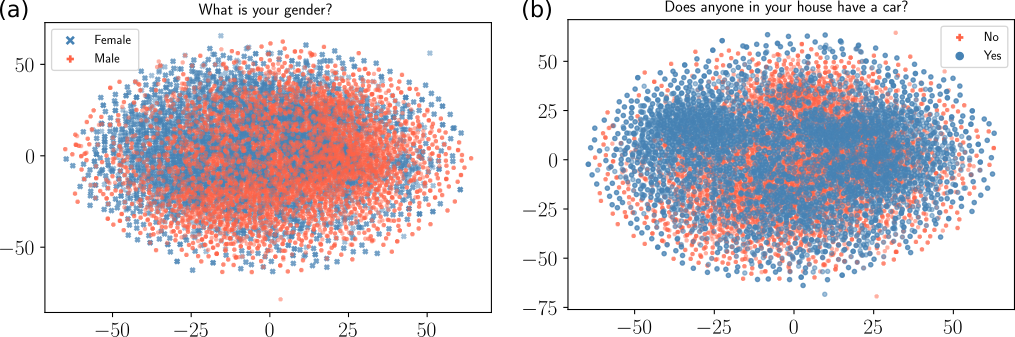}
	\caption{The above picture presents a t-sne feature embedding for a sample of the  PeNSE survey. The embedding was generated removing the physical activities questions. The distances between each instance (row) was obtained using the cosine similarity without
	any kind of weighting. The Fig. (a) and (b) was colored using the answers for the two questions appearing in the top of the pictures.}
	\label{figMultiLevelGenderAndCarNoCentralityScore}
	\end{figure}

\figref{figMultiLevelGenderAndCarNoCentralityScore} represents the embedding of each row of the survey got by the t-SNE where the colors represent a specific answer for the question above the picture. To compute the t-SNE we used the cosine similarity between each row. Notice the absence of any clustering formation in the embedding space.

\figref{fig:Diagram}

\begin{figure}[ht]
	\centering
	\includegraphics[width=.75\columnwidth]
	{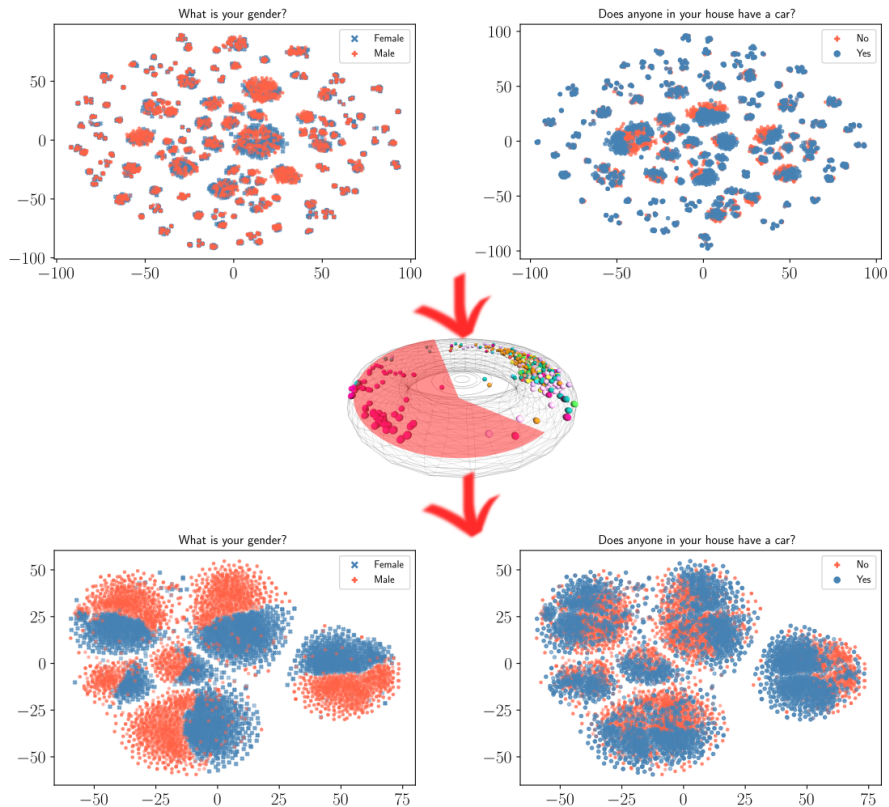}

	\caption{Circular This image represents two different embeddings of the rows related to the PeNSE survey colored by the answers to two different questions: gender and the possession of a car by someone in the family. Meanwhile, the pictures on the top shows the UMAP embedding weighted by the hub score of the interpretability graph with all the questions. The two pictures on the bottom shows the embedding result regarding a subset of questions (removing the questions marked as red in the middle of the figure) got without the phstical activities questions.}
	\label{figMultiLevelGenderAndCar}
\end{figure}

To improve the embedding result from t-sne or UMAP one solution is to perform a weighting in the features using some quantity. This technique is well known and recently was used again in the work \cite{relevanceAggregation2021} using as weight the feature importance. Here, we used a centrality measure derived from the interpretability graph, the hub score, to define this weight. Unfortunately, as a shown in the two pictures on the top of \figref{figMultiLevelGenderAndCar} we have a huge number of small clusters present in the space. But, removing somes questions based on the toroidal space obtained from the interpretability graph to segment our features into two groups are performing again the embedding technique we obtained well defined clusters regarding the gender question.

\section{Conclusions}
\label{sec:conclusions}

Graphs have largely been used to study artificial and real systems, mainly thanks to its direct formalism on modelling relationships. Knowledge in this field has proven to be useful in the study of a variety of problem and data.  In this work we report a method that uses recent developments in community and spectral analysis of graphs as well as machine learning interpretability to unravel relationships of the features in a tabular dataset. The proposed method differ from related works mainly by: (1) providing the possibility of interpreting the importance of features in predicting each other and, (2) allowing the study of the data to focus on each observation or to encompass the entire dataset.

To perform the graph analysis proposed in this work it was necessary first to develop a method to map a tabular dataset into a graph that avoids the issues present in previous works. In this method, the graph is modelled having features as vertices and the importance of each feature in predicting another as the weight value of the corresponding directed edge. These weights are assigned by considering the SHAP values of respective predictions of a machine learning model. Since the edges weights are computed for each pair of features, the resulting graph is complete. The complexity of this structure restricts the scope for graph analysis methods that can be effectively applied to it. Therefore, the disparity filter criterion was employed to keep just the strong relationships. From the filtered graph, we showed how to use  graph analysis methods to extract insights and improve the understanding of the dataset. Specifically, we analyze the  the toroidal embedding obtained by the  magnetic Laplacian and the nested stochastic block model to unravel how the features of the dataset group into communities.
The resulting modular structure, in turn, allows us to analyze the groups according to varying levels of granularity, thanks to its hierarchical grouping capabilities.

The usefulness of our methodology is exemplified respectively to the PeNSE survey dataset. The results showed several findings such as the good overall agreement between the communities obtained and the original qualitative classification of the questions in the survey, especially for the groups \textit{Food} and \textit{Body Image}.  However, the method also showed that some questions from the class \textit{Safety} could also be reassigned as \textit{Drug Consumption} questions. Also, a high connectivity was observed between the questions from the class \textit{Food} related to eating with parents and questions from the class \textit{Situations at Home}, maybe  related to the harmonious relationship with the parents.

It is important to understand the scope and limitations of the proposed approach while aiming at developing future works. For instance, the obtained graph takes into account the predictions of a machine learning model, but it does not aim at representing a causality graph. Stronger conditions would need to be satisfied to construct such a graph. Also, if the data is composed of few instances, the findings may result strongly biased.  As a future work, different tabular datasets  like medical, economical, and technical could be considered. We believe also that future works could investigate the use of some synthetic models for generating extensive tabular therefore allowing more systematic investigations of the suggested method in the spectral space. In doing so, it could be possible to establish a more direct connection between the eigenvalues and eigenvectors behavior and the structural dependencies of features. 


\section*{Acknowledgments}
The authors thank CNPq (grant 307085/2018-0), CAPES and FAPESP (grants 2019/01077-3 and 15/22308-2) for financial support.
The authors thank Joao Ricardo Sato, Suzana de S. Santos, Filipi N. Silva and Thomas Peron and  for all suggestions and useful discussions.

\ifdefined\pdfforpub
\bibliographystyle{plain}
\else
\fi

\bibliography{Projetos_tabular2vec.bib}

\begin{thebibliography}{10}

\bibitem{adamsSiriusMutualInformation2021}
Jane~L. Adams, Todd~F. Deluca, Christopher~M. Danforth, Peter~S. Dodds, Yuhang
  Zheng, Konstantinos Anastasakis, Boyoon Choi, Allison Min, and Michael~M.
  Bessey.
\newblock Sirius: {{A Mutual Information Tool}} for {{Exploratory
  Visualization}} of {{Mixed Data}}.
\newblock {\em arXiv:2106.05260 [cs, stat]}, June 2021.

\bibitem{ambriolaokuPotentialConfoundersAnalysis2019}
Amanda~Yumi Ambriola~Oku, Guilherme~Augusto Zimeo~Morais, Ana~Paula
  Arantes~Bueno, Andr{\'e} Fujita, and Jo{\~a}o~Ricardo Sato.
\newblock Potential {{Confounders}} in the {{Analysis}} of {{Brazilian
  Adolescent}}'s {{Health}}: {{A Combination}} of {{Machine Learning}} and
  {{Graph Theory}}.
\newblock {\em International Journal of Environmental Research and Public
  Health}, 17(1):90, December 2019.

\bibitem{pense2014asthma}
Maurício~Lima Barreto, Rita de~Cássia Ribeiro-Silva, Deborah~Carvalho Malta,
  Maryane Oliveira-Campos, Marco~Antonio Andreazzi, and Alvaro~Augusto Cruz.
\newblock {Prevalence of asthma symptoms among adolescents in Brazil: National
  Adolescent School-based Health Survey (PeNSE 2012)}.
\newblock {\em Revista Brasileira de Epidemiologia}, 17(1):106–115, 2014.

\bibitem{batsonSpectralSparsificationGraphs2013}
Joshua Batson, Daniel~A. Spielman, Nikhil Srivastava, and Shang-Hua Teng.
\newblock Spectral sparsification of graphs: {{Theory}} and algorithms.
\newblock {\em Communications of the ACM}, 56(8):87--94, August 2013.

\bibitem{spectraldatascience}
Yuxin Chen, Yuejie Chi, Jianqing Fan, and Cong Ma.
\newblock Spectral methods for data science: A statistical perspective.
\newblock {\em arXiv:2012.08496 [cs, eess, math, stat]}, Dec 2020.
\newblock arXiv: 2012.08496.

\bibitem{cosciaAtlasAspiringNetwork2021}
Michele Coscia.
\newblock The {{Atlas}} for the {{Aspiring Network Scientist}}.
\newblock {\em arXiv:2101.00863 [physics]}, February 2021.

\bibitem{currie2008inequalities}
Candace Currie.
\newblock {\em Inequalities in young people's health: HBSC international report
  from the 2005/2006 Survey}.
\newblock Number~5 in 1. World Health Organization, 2008.

\bibitem{da2008multiple}
Ant{\^o}nio N{\'e}lson~Rodrigues Da~Silva, Marcela da~Silva~Costa, and
  M{\'a}rcia~Helena Macedo.
\newblock Multiple views of sustainable urban mobility: The case of brazil.
\newblock {\em Transport Policy}, 15(6):350--360, 2008.

\bibitem{oliveira2017characteristics}
Max~Moura de~Oliveira, Maryane~Oliveira Campos, Marco Antonio~Ratzsch
  de~Andreazzi, and Deborah~Carvalho Malta.
\newblock Characteristics of the {{National}} adolescent school-based health
  survey-{{PeNSE}}, {{Brazil}}.
\newblock {\em Epidemiologia e Servi\c{c}os de Sa\'ude}, 26:605--616, 2017.

\bibitem{f.deresendeCharacterizationComparisonLarge2020}
Bruno~Messias {F. de Resende} and Luciano {da F. Costa}.
\newblock Characterization and comparison of large directed networks through
  the spectra of the magnetic {{Laplacian}}.
\newblock {\em Chaos: An Interdisciplinary Journal of Nonlinear Science},
  30(7):073141, July 2020.

\bibitem{fanuelDeformedLaplaciansSpectral2019}
M.~Fanuel and J.A.K. Suykens.
\newblock Deformed {{Laplacians}} and spectral ranking in directed networks.
\newblock {\em Applied and Computational Harmonic Analysis}, 47(2):397--422,
  September 2019.

\bibitem{fanuelMagneticEigenmapsVisualization2018}
Micha{\"e}l Fanuel, Carlos~M. Ala{\'i}z, {\'A}ngela Fern{\'a}ndez, and
  Johan~A.K. Suykens.
\newblock Magnetic {{Eigenmaps}} for the visualization of directed networks.
\newblock {\em Applied and Computational Harmonic Analysis}, 44(1):189--199,
  January 2018.

\bibitem{sbmTopicModelScience}
Martin Gerlach, Tiago~P. Peixoto, and Eduardo~G. Altmann.
\newblock A network approach to topic models.
\newblock {\em Science Advances}, 4(7):eaaq1360, 2017.

\bibitem{ghoshComprehensiveReviewTools2018}
Aindrila Ghosh, Mona Nashaat, James Miller, Shaikh Quader, and Chad Marston.
\newblock A comprehensive review of tools for exploratory analysis of tabular
  industrial datasets.
\newblock {\em Visual Informatics}, 2(4):235--253, December 2018.

\bibitem{relevanceAggregation2021}
Bruno~Iochins Grisci, Mathias~J. Krause, and Marcio Dorn.
\newblock Relevance aggregation for neural networks interpretability and
  knowledge discovery on tabular data.
\newblock {\em Information Sciences}, 559:111–129, Jun 2021.

\bibitem{node2vec2016}
Aditya Grover and Jure Leskovec.
\newblock node2vec: Scalable feature learning for networks.
\newblock In {\em Proceedings of the 22nd ACM SIGKDD International Conference
  on Knowledge Discovery and Data Mining}, page 855–864. ACM, Aug 2016.

\bibitem{grunbaumYouthRiskBehavior2004}
Jo~Anne Grunbaum, Laura Kann, Steve Kinchen, James Ross, Joseph Hawkins,
  Richard Lowry, William~A. Harris, Tim McManus, David Chyen, and Janet
  Collins.
\newblock Youth {{Risk Behavior Surveillance}} - {{United States}}, 2003
  ({{Abridged}}).
\newblock {\em Journal of School Health}, 74(8):307--324, October 2004.

\bibitem{guimera2004}
Roger Guimerà, Marta Sales-Pardo, and Luís A.~Nunes Amaral.
\newblock Modularity from fluctuations in random graphs and complex networks.
\newblock {\em Physical Review E}, 70(2):025101, Aug 2004.

\bibitem{pense2014drugs}
Rogério~Lessa Horta, Bernardo~Lessa Horta, Andre Wallace Nery~da Costa,
  Rogério Ruscitto~do Prado, Maryane Oliveira-Campos, and Deborah~Carvalho
  Malta.
\newblock {Lifetime use of illicit drugs and associated factors among Brazilian
  schoolchildren, National Adolescent School-based Health Survey (PeNSE 2012)}.
\newblock {\em Revista Brasileira de Epidemiologia}, 17(1):31–45, 2014.

\bibitem{tsneART}
Dmitry Kobak and Philipp Berens.
\newblock The art of using t-sne for single-cell transcriptomics.
\newblock {\em Nature Communications}, 10(11):5416, Nov 2019.

\bibitem{kuhnContributionsTheoryGames1953}
Harold~William Kuhn and Albert~William Tucker, editors.
\newblock {\em Contributions to the {{Theory}} of {{Games}} ({{AM}}-28),
  {{Volume II}}}.
\newblock {Princeton University Press}, December 1953.

\bibitem{landauZurRelativenWertbemessung1895}
Edmnund Landau.
\newblock Zur relativen wertbemessung der turnierresultate.
\newblock {\em Deutsches Wochenschach}, 11:366--369, 1895.

\bibitem{landauUberPreisverteilungBei1915}
Edmund Landau.
\newblock \"uber {{Preisverteilung}} bei {{Spieltunrnieren}}.
\newblock {\em Zeitschrift f\"ur Mathematik und Physik}, 64, 1915.

\bibitem{correlatioNet2008}
Peter Langfelder and Steve Horvath.
\newblock Wgcna: an r package for weighted correlation network analysis.
\newblock {\em BMC Bioinformatics}, 9(1):559, Dec 2008.

\bibitem{levy2010consumo}
Renata~Bertazzi Levy, In{\^e}s Rugani~Ribeiro de~Castro, Let{\'i}cia
  de~Oliveira Cardoso, Let{\'i}cia~Ferreira Tavares, Luciana
  Monteiro~Vasconcelos Sardinha, Fabio da~Silva Gomes, and Andr{\'e}
  Wallace~Nery da~Costa.
\newblock Consumo e comportamento alimentar entre adolescentes brasileiros:
  {{Pesquisa Nacional}} de {{Sa\'ude}} do {{Escolar}} ({{PeNSE}}), 2009.
\newblock {\em Ci\^encia \& Sa\'ude Coletiva}, 15:3085--3097, 2010.

\bibitem{Li_Yuan_Wu_Lu_2018}
Yuemeng Li, Shuhan Yuan, Xintao Wu, and Aidong Lu.
\newblock On spectral analysis of directed signed graphs.
\newblock {\em International Journal of Data Science and Analytics},
  6(2):147–162, Sep 2018.

\bibitem{lundbergLocalExplanationsGlobal2020}
Scott~M. Lundberg, Gabriel Erion, Hugh Chen, Alex DeGrave, Jordan~M. Prutkin,
  Bala Nair, Ronit Katz, Jonathan Himmelfarb, Nisha Bansal, and Su-In Lee.
\newblock From local explanations to global understanding with explainable
  {{AI}} for trees.
\newblock {\em Nature Machine Intelligence}, 2(1):56--67, January 2020.

\bibitem{shapConsistency2019}
Scott~M. Lundberg, Gabriel~G. Erion, and Su-In Lee.
\newblock Consistent individualized feature attribution for tree ensembles.
\newblock {\em arXiv:1802.03888 [cs, stat]}, Mar 2019.
\newblock arXiv: 1802.03888.

\bibitem{lundbergUnifiedApproachInterpreting2017}
Scott~M. Lundberg and Su-In Lee.
\newblock A unified approach to interpreting model predictions.
\newblock In {\em Proceedings of the 31st {{International Conference}} on
  {{Neural Information Processing Systems}}}, {{NIPS}}'17, pages 4768--4777,
  {Red Hook, NY, USA}, December 2017. {Curran Associates Inc.}

\bibitem{maltaTrendRiskProtective2014b}
Deborah~Carvalho Malta, Marco Antonio~Ratzsch de~Andreazzi, Maryane
  {Oliveira-Campos}, Silvania Suely Carib{\'e} de~Ara{\'u}jo Andrade, Na{\'i}za
  Nayla~Bandeira de~S{\'a}, Lenildo de~Moura, Antonio Jos{\'e}~Ribeiro Dias,
  Claudio~Dutra Crespo, and Jarbas~Barbosa da~Silva~J{\'u}nior.
\newblock Trend of the risk and protective factors of chronic diseases in
  adolescents, {{National Adolescent School}}-based {{Health Survey}}
  ({{PeNSE}} 2009 e 2012).
\newblock {\em Revista Brasileira de Epidemiologia}, 17(1):77--91, 2014.

\bibitem{pense2014drugs2}
Deborah~Carvalho Malta, Maryane Oliveira-Campos, Rogério Ruscitto~do Prado,
  Silvania Suely~Caribé Andrade, Flávia Carvalho Malta~de Mello, Antonio
  José~Ribeiro Dias, and Denise~Birche Bomtempo.
\newblock {Psychoactive substance use, family context and mental health among
  Brazilian adolescents, National Adolescent School-based Health Survey (PeNSE
  2012)}.
\newblock {\em Revista Brasileira de Epidemiologia}, 17(1):46–61, 2014.

\bibitem{maltaBullyingBrazilianSchools2010}
Deborah~Carvalho Malta, Marta Ang{\'e}lica~Iossi Silva, Flavia Carvalho~Malta
  de~Mello, Rosane~Aparecida Monteiro, Luciana Monteiro~Vasconcelos Sardinha,
  Claudio Crespo, M{\'e}rcia Gomes~Oliveira de~Carvalho, Marta Maria~Alves
  da~Silva, and Denise~Lopes Porto.
\newblock Bullying in {{Brazilian}} schools: Results from the {{National
  School}}-based {{Health Survey}} ({{PeNSE}}), 2009.
\newblock {\em Ci\^encia \& Sa\'ude Coletiva}, 15:3065--3076, October 2010.

\bibitem{marcaccioliPolyaUrnApproach2019}
Riccardo Marcaccioli and Giacomo Livan.
\newblock A {{P\'olya}} urn approach to information filtering in complex
  networks.
\newblock {\em Nature Communications}, 10(1):745, December 2019.

\bibitem{umapPaper}
Leland McInnes, John Healy, and James Melville.
\newblock Umap: Uniform manifold approximation and projection for dimension
  reduction.
\newblock (arXiv:1802.03426), Sep 2020.
\newblock arXiv:1802.03426 [cs, stat].

\bibitem{molnarInterpretableMachineLearning}
Christoph Molnar.
\newblock Interpretable {{Machine Learning}}, 2019.

\bibitem{ganho1987}
Dan~H. Moore.
\newblock Classification and regression trees.
\newblock {\em Cytometry}, 8(5):534–535, Sep 1987.

\bibitem{nsbm2021}
Leonardo Morelli, Valentina Giansanti, and Davide Cittaro.
\newblock Nested stochastic block models applied to the analysis of single cell
  data.
\newblock {\em {bioRxiv}}, page 2020.06.28.176180, Apr 2021.

\bibitem{gradBoostTutorial2013}
Alexey Natekin and Alois Knoll.
\newblock Gradient boosting machines, a tutorial.
\newblock {\em Frontiers in Neurorobotics}, 7:21, 2013.

\bibitem{Newman_Girvan_2004}
M.~E.~J. Newman and M.~Girvan.
\newblock Finding and evaluating community structure in networks.
\newblock {\em Physical Review E}, 69(2):026113, Feb 2004.

\bibitem{geneNetwork2019}
Magdalena Niemira, Francois Collin, Anna Szalkowska, Agnieszka Bielska,
  Karolina Chwialkowska, Joanna Reszec, Jacek Niklinski, Miroslaw Kwasniewski,
  and Adam Kretowski.
\newblock Molecular signature of subtypes of non-small-cell lung cancer by
  large-scale transcriptional profiling: Identification of key modules and
  genes by weighted gene co-expression network analysis (wgcna).
\newblock {\em Cancers}, 12(1):37, Dec 2019.

\bibitem{peixotoHierarchicalBlockStructures2014}
Tiago~P. Peixoto.
\newblock Hierarchical {{Block Structures}} and {{High}}-{{Resolution Model
  Selection}} in {{Large Networks}}.
\newblock {\em Physical Review X}, 4(1):011047, March 2014.

\bibitem{peixotoNonparametricBayesianInference2017}
Tiago~P Peixoto.
\newblock Nonparametric bayesian inference of the microcanonical stochastic
  block model.
\newblock {\em Physical Review E}, 95(1):012317, 2017.

\bibitem{sbmNsbm2020}
Tiago~P. Peixoto.
\newblock Merge-split markov chain monte carlo for community detection.
\newblock {\em Physical Review E}, 102(1):012305, Jul 2020.

\bibitem{samekLearningExplainableTrees2020}
Wojciech Samek.
\newblock Learning with explainable trees.
\newblock {\em Nature Machine Intelligence}, 2(1):16--17, January 2020.

\bibitem{serranoExtractingMultiscaleBackbone2009}
M.~A. Serrano, M.~Boguna, and A.~Vespignani.
\newblock Extracting the multiscale backbone of complex weighted networks.
\newblock {\em Proceedings of the National Academy of Sciences},
  106(16):6483--6488, April 2009.

\bibitem{imageSegSpectra}
Jianbo Shi and J.~Malik.
\newblock Normalized cuts and image segmentation.
\newblock In {\em Proceedings of IEEE Computer Society Conference on Computer
  Vision and Pattern Recognition}, page 731–737. IEEE Comput. Soc, 1997.

\bibitem{shubinDiscreteMagneticLaplacian1994a}
M.~A. Shubin.
\newblock Discrete {{Magnetic Laplacian}}.
\newblock {\em Communications in Mathematical Physics}, 164(2):259--275, August
  1994.

\bibitem{pense2020sedentary}
Roberta Mendes~Abreu Silva, Amanda Cristina de~Souza Andrade, Waleska~Teixeira
  Caiaffa, Danielle Souto~de Medeiros, and Vanessa~Moraes Bezerra.
\newblock National adolescent school-based health survey - pense 2015:
  Sedentary behavior and its correlates.
\newblock {\em PLOS ONE}, 15(1):e0228373, Jan 2020.

\bibitem{vignaSpectralRanking2019}
Sebastiano Vigna.
\newblock Spectral {{Ranking}}.
\newblock {\em arXiv:0912.0238 [physics]}, February 2019.

\bibitem{magnet2021}
Xitong Zhang, Yixuan He, Nathan Brugnone, Michael Perlmutter, and Matthew Hirn.
\newblock Magnet: A neural network for directed graphs.
\newblock {\em arXiv:2102.11391 [cs]}, Jun 2021.
\newblock arXiv: 2102.11391.

\end{thebibliography}
\appendix
\end{document}